# Mechanics of compliant serial manipulator composed of dual-triangle segments

*Wanda Zhao*
Laboratoire des Sciences du Numérique de Nantes
(LS2N), UMR CNRS 6004
Ecole Centrale Nantes
Nantes, France
Wanda.Zhao@ls2n.fr

*Anatol Pashkevich*
Laboratoire des Sciences du Numérique de Nantes
(LS2N), UMR CNRS 6004
IMT Atlantique Nantes
Nantes, France
Anatol.Pashkevich@imt-atlantique.fr

*Alexandr Klimchik*
Center Technologies in Robotics and Mechatronics
Components
Innopolis University
Tatarstan, Russia
A.Klimchik@innopolis.ru

*Damien Chablat*
Laboratoire des Sciences du Numérique de Nantes
(LS2N), UMR CNRS 6004
Centre National de la Recherche Scientifique (CNRS)
Nantes, France
Damien.Chablat@cnrs.fr

*Abstract*—The paper focuses on the mechanics of a compliant serial manipulator composed of new type of dual-triangle elastic segments. Both the analytical and numerical methods were used to find the manipulator stable and unstable equilibrium configurations, as well as to predict corresponding manipulator shapes. The stiffness analysis was carried on for both loaded and unloaded modes, the stiffness matrices were computed using the Virtual Joint Method (VJM). The results demonstrate that either buckling or quasi-buckling phenomenon may occur under the loading, if the manipulator corresponding initial configuration is straight or non-straight one. Relevant simulation results are presented that confirm the theoretical study.

## I. INTRODUCTION

Currently, compliant serial manipulators are used more and more in many applications (such as inspection in constraint environment, medical fields etc.), because of their sophisticated motions and low weight. Conventional compliant manipulators are usually composed of rigid links and compliant actuators, like hinges, axles, or bearings. However, there is a growing number of research in this area dealing with some new mechanical structures [1][2][3][4], which allow to implement compliant motions by means of tensegrity mechanisms, one of which is studied in this paper.

In general, the robotic manipulators are usually classified into three types [5], conventional discrete, serpentine, and continuum robots. The first one is made of traditional rigid components. The second one use discrete joints but combine very short rigid links with a large density joints, which produce smooth curves and make the robot similar to a snake or elephant trunk [6]. While the continuum robots do not contain any rigid links or joints, they are very smooth and soft, bending continuously when working [7]. Many researchers have done studies on serpentine and continuum robots in recent years, designed flexible mechanisms for many applications [8]. However, the pure soft continuum robot received less attention, as its small output force and difficulty of design. Thus, combining rigid and elastic or soft components becomes a popular practice in designing robot manipulator. The typical earlier hyper-redundant robot designs and implementations can be date to 1970's [9], which includes a series of plates interconnected by universal joint and elastic control components for pivotable action with respect to one another. [10][11][12][13][14]

Nowadays, a very promising trend in compliant robotics is using a series of similar segments based on varies tensegrity mechanisms, which are assembly of compressive elements and tensile elements (cables or springs) held together in equilibrium [15][16]. Some kinds of the tensegrity mechanisms have been already studied carefully. Such as [17], the authors deal with the mechanism composed of two springs and two length-changeable bars. They analyzed the mechanism stiffness using the energy method, demonstrated that the mechanism stiffness always decreasing under external loading with the actuators locked, which may lead to "buckling". And in [18][19], the cable-driven X-shape tensegrity structures were considered, the authors investigated the influence of cable lengths on the mechanism equilibrium configurations, which maybe both stable and unstable. The relevant analysis of equilibrium configurations stability and singularity can be seen in [20].

A new type of compliant tensegrity mechanism was proposed in our previous papers [21][22]. It is composed of two rigid triangle parts, which are connected by a passive joint in the center and two elastic edges on each sides with controllable preload. The stiffness analysis of a basic dual-triangle was carried on, and the stable condition of the equilibrium was obtained. The results also shown that maybe the buckling phenomenon here. Usually, while designing a robot, researchers always try to avoid the buckling, but such behavior can make improvements in some fields [22]. So this phenomenon must be taken into account by the designer of serial manipulators. In this paper, we study a compliant serial manipulator composed of the dual-triangle segments mentioned above, and concentrate on the equilibrium configurations and their transformations under the loading that may be either continuous or sporadic, leading to buckling phenomenon. Both the loaded and unloaded stiffness model of this manipulator were analyzed, and the simulation of the manipulator behavior after buckling was obtained, which provide a good base for such manipulators mechanical design and development of relevant control algorisms.

## II. MECHANICS OF DUAL-TRIANGLE MECHANISM

Let us consider first a single segment of the total serial manipulator to be studied, which consists of two rigid triangles connected by a passive joint whose rotation is constrained by two linear springs as shown in Fig. 1. It is assumed that the mechanism geometry is described by the triangle parameters $(a_1, b_1)$ and $(a_2, b_2)$, and the mechanism shape is defined by the angle that can be adjusted by means of two control inputs influencing on the spring lengths $L_1$ and $L_2$. Let us denote the spring lengths in the non-stress state as $L_1^0$ and $L_2^0$, and the spring stiffness coefficients $k_1$ and $k_2$.

To find the mechanism configuration angle $q$ corresponding to the given control inputs $L_1^0$ and $L_2^0$, let us derive first the static equilibrium equation. From Hook's law, the forces generated by the springs are $F_i = k_i(L_i - L_i^0)$ ($i=1, 2$), where $L_1$ and $L_2$ are the spring lengths |AD|, |BC| corresponding to the current value of the angle $q$. These values can be computed using the formulas $L_i(\theta_i) = \sqrt{c_1^2 + c_2^2 + 2c_1 c_2 \cos(\theta_i)}$ ($i=1,2$), here $c_i = \sqrt{a_i^2 + b_i^2}$ ($i=1,2$), and the angles $\theta_1$, $\theta_2$ are expressed via the mechanism parameters as $\theta_1 = \beta_{12} + q$, $\theta_2 = \beta_{12} - q$, $\beta_{12} = \operatorname{atan}(a_1/b_1) + \operatorname{atan}(a_2/b_2)$. The torques $M_1 = F_1 \cdot h_1$, $M_1 = F_2 \cdot h_2$ in the passive joint O can be computed from the geometry, so we can get

$$M_1(q) = +k_1(1 - L_1^0/L_1(\theta_1))\, c_1 c_2 \sin(\theta_1)$$
$$M_2(q) = -k_2(1 - L_2^0/L_2(\theta_2))\, c_1 c_2 \sin(\theta_2) \quad (1)$$

where the difference in signs is caused by the different direction of the torques generated by the forces $F_1$, $F_2$ with respect to the passive joint. Further, taking into account the external torque $M_{\text{ext}}$ applied to the moving platform, the static equilibrium equation for the considered mechanism can be written as $M_1(q) + M_2(q) + M_{\text{ext}} = 0$.

Let us now evaluate the stability of the mechanism under consideration. In general, this property highly depends on the equilibrium configuration defined by the angle $q$, which satisfies the equilibrium equation $M(q) + M_{\text{ext}} = 0$. As follows from the relevant analysis, the function $M(q)$ can be either monotonic or non-monotonic one, so the single-segment mechanism may have multiple stable and unstable equilibriums, which are studied in detail [21][21]. As Fig. 2 shows, the torque-angle curves $M(q)$ that can be either monotonic or two-model one, the considered stability condition can be simplified and reduced to the derivative sign verification at the zero point only, i.e.

$$M'(q)\big|_{q=0} < 0, \quad (2)$$

which is easy to verify in practice. It represents the mechanism equivalent rotational stiffness for unloaded configuration with $q=0$.

Let us also consider in detail the symmetrical case, for which $a_1=a_2=a$, $b_1=b_2=b$, $k_1=k_2$, $L_i^0 = L^0$. Then as follows from the mechanism geometry, to distinguish the monotonic and non-monotonic cases presented in Fig. 2, we can omit some indices and present the torque-angle relationship as well as the stiffness expression in the more compact forms:

$$M(q) = 2ck\left[c \cos\beta_{12} \sin q - L^0 \cos(\beta_{12}/2)\sin(q/2)\right]$$
$$M'(q) = ck\left[2c \cos\beta_{12} \cos q - L^0 \cos(\beta_{12}/2)\cos(q/2)\right], \quad (3)$$

it is also necessary to compute $M'(q)$ for unloaded equilibrium configuration $q=0$, which allows us to present the condition of torque-angle curve monotonicity as follows

$$L^0 > 2b \cdot \left(1 - (a/b)^2\right), \quad (4)$$

that will be used in further analysis.

## III. MECHANICS OF SERIAL MANIPULATOR

### A. Manipulator Geometry and Kinematics

Let us consider a manipulator composed of three similar segments connected in series as shown in Fig. 3, where the left hand-side is fixed and the initial configuration is a "straight" one ($q_1=q_2=q_3=0$). This configuration is achieved by applying equal control inputs to all mechanism segments. For this manipulator, it is necessary to investigate the influence of the external force $\mathbf{F}_e=(F_x, F_y)$, which causes the end-effector displacements to a new equilibrium location $(x, y)^T = (6b - \delta_x, \delta_y)^T$ corresponding to the nonzero configuration variables ($q_1, q_2, q_3$). It is also assumed here the external torque $M_{\text{ext}}$ applied to the end-effector is equal to zero. It can be easily proved from the geometry analysis that the configuration angles satisfy the following direct kinematic equations

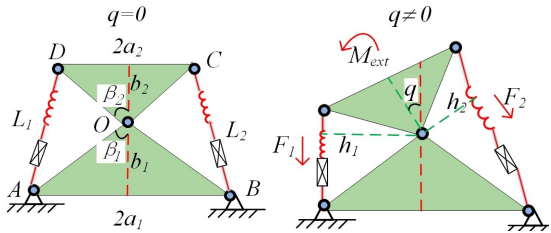

Figure 1. Geometry of a single dual-triangle mechanism.

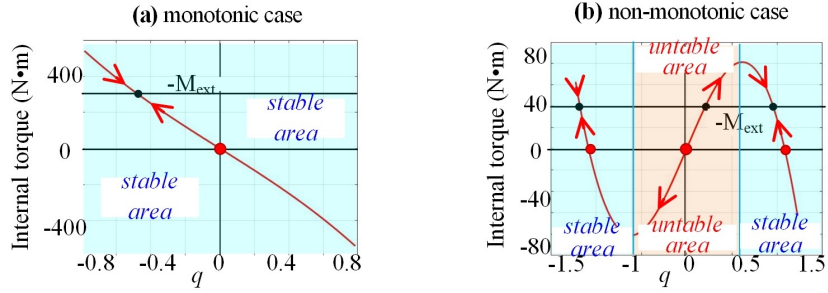

Figure 2. The torque-angle curves and static equilibriums for $L_1^0 = L_2^0$ ($q_0 = 0$).

$$\begin{aligned} x &= b + 2bC_1 + 2bC_{12} + bC_{123} \\ y &= 2bS_1 + 2bS_{12} + bS_{123} \end{aligned}, \quad (5)$$

where $C_{123} = \cos(q_1 + q_2 + q_3)$, $S_{123} = \sin(q_1 + q_2 + q_3)$, $C_{12} = \cos(q_1 + q_2)$, $S_{12} = \sin(q_1 + q_2)$, $C_1 = \cos q_1$, $S_1 = \sin q_1$. These two equations include three unknown variables and allow us to compute two of them assuming that the remaining one is known. For instance, if the angle $q_1$ is assumed to be known, the rest of the angles $q_2$, $q_3$ can be computed from the classical invers kinematics of the two-link manipulator as follows

$$\begin{aligned} q_3 &= \operatorname{atan}(S_3/C_3) \\ q_2 &= \operatorname{atan}(y - 2bS_1/x - b - 2bC_1) - \operatorname{atan}(bS_3/2b + bC_3) - q_1 \end{aligned} \quad (6)$$

where $C_3 = \left[(x - b - 2bC_1)^2 + (y - 2bS_1)^2 - 5b^2\right]/4b^2$, $S_3 = \pm\sqrt{1 - C_3^2}$. It is clear that the latter expressions provides two group of possible solutions corresponding to the positive/negative configuration angles $q_3 \geq 0$ and $q_3 \leq 0$.

To find a stable manipulator configuration under the loading, let us apply the energy method. It is clear that the end-effector displacement caused by the external loading leads to some deflections in the mechanism springs, which allows us to compute the manipulator energy as

$$E = \frac{1}{2}\sum_{i=1}^{3}\sum_{j=1}^{2} k\left(L_{ij} - L_{ij}^0\right)^2, \quad (7)$$

where $L_{ij}$ and $L_{ij}^o$ are the spring lengths in current and initial (unextended) states respectively. Because the manipulator end-effector is assumed to be fixed at the point, the above energy can be expressed via one of the three variables $q_1$, $q_2$ or $q_3$. Assuming that variable $q_1$ is chosen as an independent one, the desired stable configurations can be found by computing local minima of energy function

$$E(q_1) \to \min_{q_1}. \quad (8)$$

Examples of such energy curves $E(q_1)$ for several typical cases are presented in Fig. 4.

### B. Manipulator Stiffness Behavior

An alternative way to compute the configuration angles $q_1$, $q_2$, $q_3$ at the equilibrium state is based on the torque equation $M_e(q_1)=0$, which is implicitly used in the energy method. The latter is illustrated by combined plots of the energy-torque curves computed for the initial "straight" configuration presented in Fig. 5, which shows that the max/min of the energy $E(q_1)$ correspond to zeros of the torque $M_e(q_1)=0$. Further, to find the external forces corresponding to this end-point location, it is necessary to use the force-torque equilibrium equation.

$$\mathbf{M} + \mathbf{J}_q^T \mathbf{F} = 0, \quad (9)$$

where $\mathbf{M}=(M_{q1}, M_{q2}, M_{q3})^T$, $\mathbf{F}=(F_x, F_y, M_e)^T$, and they relates internal torques $M_{q1}$, $M_{q2}$ and $M_{q3}$ in all manipulator segments and the force/torque at the end-point. In this equation, the internal torques can be computed using previously derived expression from section II,

$$M_{qi} = 2k\left[(b^2 - a^2)\cdot\sin q_i - bL^0 \sin(0.5q_i)\right]; \quad i = 1, 2, 3, \quad (10)$$

and the Jacobian matrix $\mathbf{J}_q$ can be computed using the standard technique for the three-link manipulator and

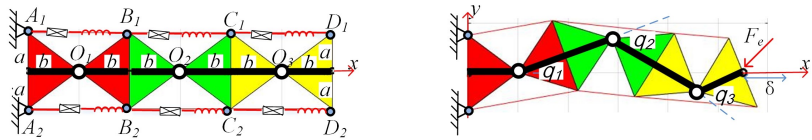

Figure 3. The torque-angle curves and static equilibriums for $L_1^0 = L_2^0$ ($q_0 = 0$).

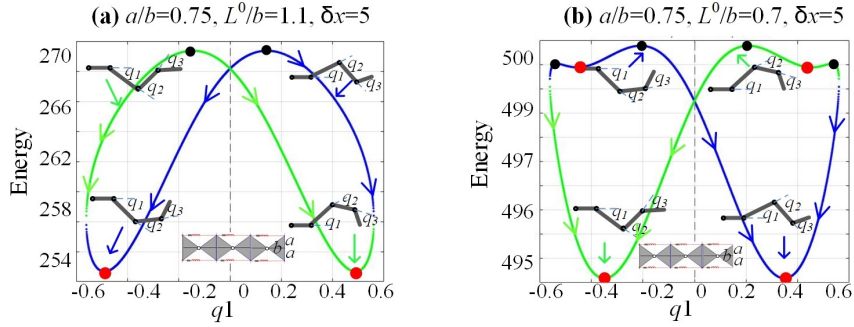

Figure 4. Energy curves $E(q_1)$ for different combinations of manipulator geometric parameters $a/b$, $L^0/b$:
"blue curves" — positive configuration with $q_3>0$; "green curves" — negative configuration with $q_3<0$;
● — stable equilibrium; ● — unstable equilibrium

presented as follows

$$\mathbf{J}_q = \begin{bmatrix} -2bS_1 - 2bS_{12} - bS_{123} & -2bS_{12} - bS_{123} & -bS_{123} \\ 2bC_1 + 2bC_{12} + bC_{123} & 2bC_{12} + bC_{123} & bC_{123} \\ 1 & 1 & 1 \end{bmatrix}, (11)$$

where $S$ and $C$ with corresponding indices have the same meaning as in (5). Assuming that the Jacobian is non-singular (i.e. the loaded manipulator is already out of the straight configuration), the external force/torque can be expressed directly as $\mathbf{F} = -\mathbf{J}_q^{-T}\mathbf{M}$, where the transport inverse matrix $\mathbf{J}_q^{-T}$ can be computed analytically, leading to the following expression

$$\begin{pmatrix} F_x \\ F_y \\ M_e \end{pmatrix} = -\frac{1}{2bS_2}\begin{bmatrix} C_{12} & -C_1 - C_{12} & C_1 \\ S_{12} & -S_1 - S_{12} & S_1 \\ bS_3 & -bS_{23} - bS_3 & 2bS_2 + bS_{23} \end{bmatrix}\begin{bmatrix} M_{q1} \\ M_{q2} \\ M_{q3} \end{bmatrix}, (12)$$

The latter allows us to rewrite the system of the equilibrium

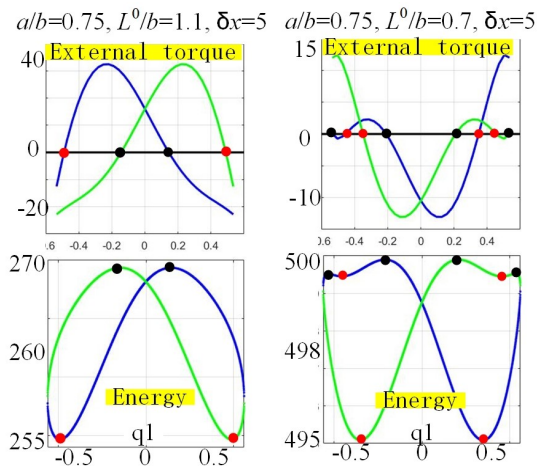

Figure 5. Correspondence between the maxima/minima of the energy curves $E(q_1)$ and zeros of the external torque $M_e(q_1)$.

equation (5) in the following extended form

$$\begin{cases} b + 2bC_1 + 2bC_{12} + bC_{123} - x = 0 \\ 2bS_1 + 2bS_{12} + bS_{123} - y = 0 \\ S_3 M_{q1} - (S_{23} + S_3) M_{q2} + (2S_2 + S_{23}) M_{q3} = 0 \end{cases}, (13)$$

whose solution ($q_1$, $q_2$, $q_3$) may correspond to either to stable or unstable equilibrium of the manipulator configuration. Then, using expressions $F_x(q_1, q_2, q_3)$ and $F_y(q_1, q_2, q_3)$ obtained from (12), one can get the external loading ($F_x$, $F_y$) corresponding to the end-effector position ($x$, $y$), which finally allows us to generate the desired force-deflection curves. Examples of such curves for several case studies are presented in Fig. 6, where it is assumed that under the loading the manipulator moves along with x-axis, i.e. $\delta x = \text{var}$, $\delta y = 0$. As follows from this figure, in the most cases (Fig. 6a) the force-deflection curves are quasi-linear but some of them they include discontinuities (jumps) and do not pass through the zero point. The latter means that the corresponding manipulator possesses very specific particularity known as the "buckling" property [23][21][22], for which the configuration angles may suddenly change their values while the external force increases gradually. Besides, in the case presented in Fig. 6b, the manipulator demonstrates the "jumping" phenomenon, because the initial (unloaded position) is unstable and the manipulator suddenly changes its shape even for extremely low loading.

To compute the critical force $F_x^0$ causing the buckling, let us assume that the configuration angles ($q_1$, $q_2$, $q_3$) are small enough but not equal to zero. This allows us to derive a linearized stiffness model in the neighborhood of $q_i=0$ ($i=1, 2, 3$). Under such assumptions, the first and second equations from (3.15) can be presented in the following form

$$\begin{aligned} \delta_x &= b(q_1^2 + q_{12}^2 + 0.5 q_{123}^2) \\ \delta_y &= 2b(q_1 + q_{12} + 0.5 q_{123}) \end{aligned}, (14)$$

which allows us to present the condition $\delta_y=0$ as $q_1+q_{12}+q_{123}/2=0$. Applying similar linearization to the third equation from (13), one can get the additional relation for the configuration angles $q_1 q_3 - q_2^2 + q_2 q_3 + q_3^2 = 0$ ensuring the equality $M_e=0$. Further, combining these two obtained relations and considering $q_2$ as an independent variable, it is possible to express $q_1$, $q_3$ in the following way $q_1 = \alpha_1 \cdot q_2$, $q_3 = \alpha_3 \cdot q_2$, where

$$\alpha_1 = -\left(\pm\sqrt{21}+11\right)/20; \quad \alpha_3 = \left(\pm\sqrt{21}-1\right)/4. \quad (15)$$

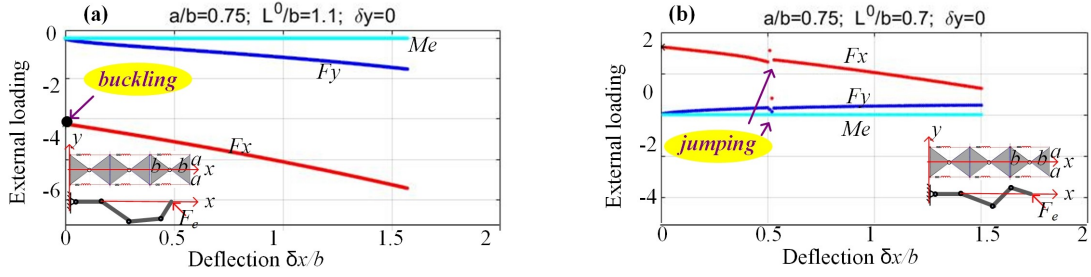

Figure 6. Force-deflection curves and stiffness coefficients for the "straight" initial configuration.

The latter gives us four possible manipulator geometric configurations corresponding to the static equilibrium, two with U-shape and two with Z-shape (see Table 1).

The corresponding external forces $F_x$, $F_y$ can be linearized for small configuration angles, which yields

$$F_x \approx -\frac{k}{2bq_2}\left[2(b^2-a^2)-bL^0\right](q_1+q_3-2q_2); \quad F_y \approx 0 \quad (16)$$

Further, taking into account (14) the desired critical force can be expressed in the following way

$$F_x^o = \lim_{q_i \to 0} F_x = -\lambda \cdot \frac{k}{b}\left[2(b^2-a^2)-bL^0\right] \quad (17)$$

where $\lambda = \left(\sqrt{21}-14\right)/10 \approx -0.9417$ for U-shape, and $\lambda = -\left(\sqrt{21}+14\right)/10 \approx -1.8583$ for Z-shape.

It is worth mentioning that the obtained expression allows also to derive the static stability condition for the straight configuration. In fact, this configuration is stable if and only if $F_x^0 < 0$, which is equivalent to $2(b^2 - a^2) < bL^0$, which defining the monotonicity of the torque-angle curves for the manipulator segments.

Finally, let us compare the U-shape and Z-shape equilibrium configurations from point of view their static stability. It can be easily proved that for the small configuration angles $q_i$ the end-effector deflection $\delta_x$ can be expressed in the following way

$$\delta_x = \mu q_2^2 \quad (18)$$

where $\mu = \left(\sqrt{21}+21\right)/20 \approx 1.2791$ for U-shape, and $\mu = \left(-\sqrt{21}+21\right)/20 \approx 0.8209$ for Z-shape. The latter means that for the similar deflections $\delta_x$, the U-shape has smaller configuration angles $q_i$ compared to the Z-shape, which ensures smaller energy in agreement with (8).

Let us consider now the case when the manipulator initial configuration is a non-straight one, which corresponds to the non-zero angles ($q_i^0 \neq 0$, $i=1,2,3$). Similar to the above section, the equilibrium is defined by three equations (13) that are derived from the direct kinematics and from the zero external torque assumption $M_e=0$. It can be also proved that here the energy curves have the "∞-shape" as for the straight configuration considered before. However, depending on the initial end-effector location $(x, y)$, these energy curves may be non-symmetrical and can be even discontinuous and include cusp points. Typical examples of such curves corresponding to end-point location $(x, y)^T = (5.5b, 0)^T$ are presented in Fig. 7 where the discontinuity caused by the geometric constraint is clearly visible. In particular, in cases (a) energy curve consists of two separate U-shape parts that yield two symmetrical stable equilibriums and four unstable ones. Such separation is caused by the geometric constrains on the angles $|q_i| \leq q_i^{max}$. However, the energy curves for the case (b) cannot be treated in the same way, because such combination of $a$, $b$, $L_i^0$ provides non-monotonic torque-angle curves for the segments, and even separate parts of the manipulator are unstable in this case. It should be stressed that in the cases (a), each segment of the mechanism is statically stable. It should be also noted that there are a number of unfeasible sections (black lines) inside all energy curves, where at least one of the angles $q_2$ or $q_3$ is out of the allowable geometric limits.

TABLE I. POSSIBLE MANIPULATOR SHAPES IN STATIC EQUILIBRIUM

|  | q1 | q2 | q3 | Geometric configuration | Stability |
|---|---|---|---|---|---|
| Case of "+√" | − | + | + | U-shape: | Stable |
|  | + | − | − | U-shape: | Stable |
| Case of "-√" | − | + | − | Z-shape: | Unstable |
|  | + | − | + | Z-shape: | Unstable |

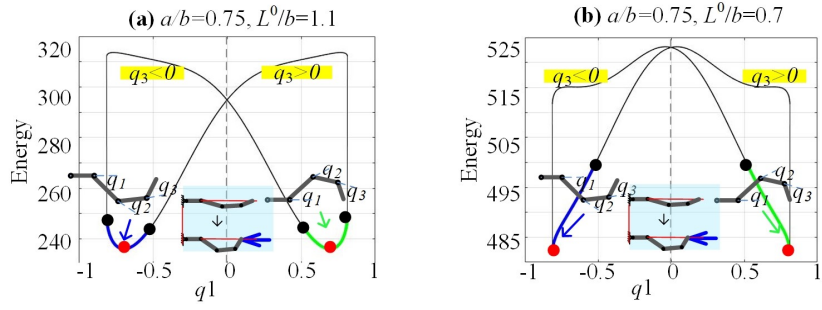

Figure 7. Energy curves $E(q_1)$ for different $(a, b, L^o)$ for non-straight initial configuration and displacement $(\Delta x, \Delta y) = (b/2, 0)$
"blue curves" — feasible configuration with $q_3>0$; "green curves" — feasible configuration with $q_3<0$;
"black curves" — unfeasible configuration; "red point ●" — stable equilibrium; "black point ●" — unstable equilibrium..

The above presented case studies corresponding to end-effector initial position $(x, y)^T = (5.5b, 0)^T$ can be also illustrated by the force-deflection curves presented in Fig. 8. As follows from them, there is no buckling phenomenon in the cases (a), the curve is quasi-linear and pass through the zero point. Besides, the buckling detected in the case (b) cannot be observed in practice because of non-stability of the separate manipulator segments.

To evaluate the manipulator stiffness matrix for the non-straight configuration, let us first find the joint torques for all manipulator segments using method from section II,

$$M_{qi} = 2k(b^2 - a^2)\sin(q_i) - kL_{i1}^0 \left[ a\cos(q_i/2) + b\sin(q_i/2) \right] \\ + kL_{i2}^0 \left[ a \cdot \cos(q_i/2) - b\sin(q_i/2) \right]; \quad i = 1, 2, 3 \quad (19)$$

and compute the derivatives providing equivalent stiffness coefficients in the joints $K_{qi} = dM_{qi}/dq_i$

$$K_{qi} = 2k(b^2 - a^2)\cos(q_i) - kL_{i1}^0 \left[ b\cos(q_i/2) - a\sin(q_i/2) \right]/2 \\ - kL_{i2}^0 \left[ a \cdot \sin(q_i/2) + b\cos(q_i/2_i) \right]/2; \quad i = 1, 2, 3 \quad (20)$$

This allows us to apply the VJM method and to express the unloaded stiffness matrix of the considered manipulator as

$$\mathbf{K}_F^0 = \left( \mathbf{J}_o \mathbf{K}_{qo}^{-1} \mathbf{J}_o^T \right)^{-1}, \quad (21)$$

where the subscript "$o$" denotes the variables corresponding to the unloaded initial configuration. Further, if we express the 2x3 submatrix of the Jacobian (11) for this configuration as

$$\mathbf{J}_o = \begin{bmatrix} J_{11} & J_{12} & J_{13} \\ J_{21} & J_{22} & J_{23} \end{bmatrix}_{2 \times 3}, \quad (22)$$

The desired compliance matrix of the unloaded mode can be expressed analytically in the following way

$$\mathbf{C}_F^0 = \mathbf{J}_o \mathbf{K}_{qo}^{-1} \mathbf{J}_o^T = \begin{bmatrix} \dfrac{J_{11}^2}{K_{q1}} + \dfrac{J_{12}^2}{K_{q2}} + \dfrac{J_{13}^2}{K_{q3}} & * \\ * & \dfrac{J_{21}^2}{K_{q1}} + \dfrac{J_{22}^2}{K_{q2}} + \dfrac{J_{23}^2}{K_{q3}} \end{bmatrix} \quad (23)$$

where $\mathbf{K}_{qo} = diag(K_{q1}, K_{q2}, K_{q3})$ is the matrix of size 3×3.

For the loaded mode, the manipulator stiffness matrix can be computed using the extended VJM technique proposed in [24]. Within this technique, let us assume that there is a non-negligible deflection $\Delta = (\Delta x, \Delta y)^T$ caused by the external force $\mathbf{F} = (F_x, F_y)^T$, and there is a small deflection $\delta = (\delta x, \delta y)^T$ caused by this force variation $\delta \mathbf{F} = (\delta F_x, \delta F_y)^T$ that corresponds to the joint angle variations $\delta q = (\delta q_1, \delta q_2, \delta q_3)^T$. As follows from the equilibrium equation $\mathbf{M} = \mathbf{J}^T \mathbf{F}$, corresponding variation of the joint torque can be expressed as

$$\delta \mathbf{M} = \left( \dfrac{d\mathbf{J}^T}{d\mathbf{q}} \delta \mathbf{q} \right) \cdot \mathbf{F} + \mathbf{J}^T \cdot \delta \mathbf{F}, \quad (24)$$

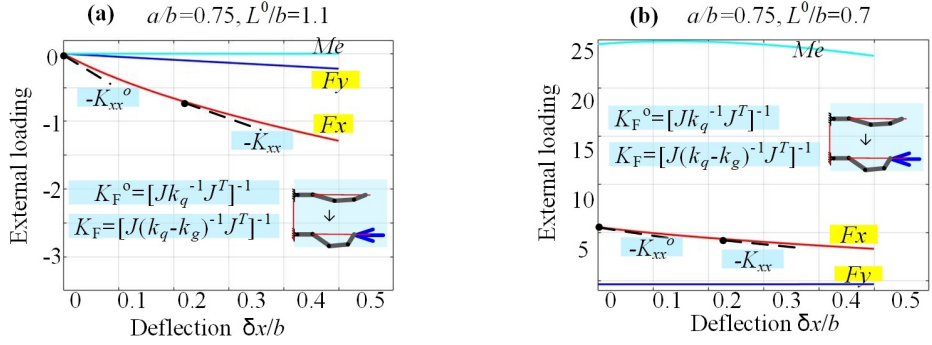

Figure 8. Force-deflection curves and stiffness coefficients for "non-straight" initial configuration
with different parameters ($a$, $b$, $L^o$) and displacement $(\Delta x, \Delta y) = (b/2, 0)$.

where the part including the Jacobian derivative $d\mathbf{J}^T/d\mathbf{q}$ can be rewritten as

$$\left(\frac{d\mathbf{J}^T}{d\mathbf{q}}\delta\mathbf{q}\right)\cdot\mathbf{F} = \sum_{i=1}^{3}\left(\frac{\partial \mathbf{J}^T}{\partial q_i}\cdot\mathbf{F}\right)\delta q_i = \mathbf{K}_g \cdot \delta\mathbf{q}, \quad (25)$$

where $\mathbf{K}_g$ is the 3×3 matrix describing the influence of loading $\mathbf{F}$ on the manipulator Jacobian $\mathbf{J}$

$$\mathbf{K}_g = \left[\frac{\partial \mathbf{J}^T}{\partial q_1}\cdot\mathbf{F} \;\middle|\; \frac{\partial \mathbf{J}^T}{\partial q_2}\cdot\mathbf{F} \;\middle|\; \frac{\partial \mathbf{J}^T}{\partial q_3}\cdot\mathbf{F}\right]_{3\times 3}, \quad (26)$$

that can be also written in the extended form as

$$\mathbf{K}_g = \begin{bmatrix} -J_{21}F_x + J_{11}F_y & -J_{22}F_x + J_{12}F_y & -J_{23}F_x + J_{13}F_y \\ -J_{22}F_x + J_{12}F_y & -J_{22}F_x + J_{12}F_y & -J_{23}F_x + J_{13}F_y \\ -J_{23}F_x + J_{13}F_y & -J_{23}F_x + J_{13}F_y & -J_{23}F_x + J_{13}F_y \end{bmatrix}_{3\times 3} \quad (27)$$

Further, after expressing the virtual joint torque variation as $\delta\mathbf{M} = \mathbf{K}_q \cdot \delta\mathbf{q}$ and its substitution to (24), the variable $\delta\mathbf{q}$ can be presented as

$$\delta\mathbf{q} = \left(\mathbf{K}_q - \mathbf{K}_g\right)^{-1}\cdot\mathbf{J}^T\cdot\delta\mathbf{F}, \quad (28)$$

which allows us to find the end-effector deflection $\boldsymbol{\delta} = \mathbf{J}\cdot\delta\mathbf{q}$ and finally to obtain the desired loaded compliance and stiffness matrices

$$\begin{aligned}\mathbf{C}_F &= \mathbf{J}\left(\mathbf{K}_q - \mathbf{K}_g\right)^{-1}\mathbf{J}^T \\ \mathbf{K}_F &= \left[\mathbf{J}\left(\mathbf{K}_q - \mathbf{K}_g\right)^{-1}\mathbf{J}^T\right]^{-1}\end{aligned}, \quad (29)$$

It is worth mentioning that all the Jacobian and the joint stiffness matrices $\mathbf{K}_q$, $\mathbf{K}_g$ must be computed for the loaded equilibrium configuration, which is different from the initial unloaded one ( It requires relevant solution of the non-linear equations considered above ).

To illustrate importance of the loaded stiffness analysis, the obtained expressions were applied to several cases study focusing on the manipulator stiffness changing under the external loading. For all considered cases, it was assumed that the initial manipulator configuration is a non-straight one, with the end point location $(x_0, y_0)=(5.5b,0)$. Under the loading the configuration angles corresponding to the external force $\mathbf{F}=(F_x, F_y)^T$ were computed from (12) numerically (using *Newton's Method*). There were compared three combinations of the geometric parameters $a/b \in \{0.75; 0.9; 1.1\}$, relevant results are presented in Figs. 9 and 10. As follows from these figures, in most cases the manipulator stiffness essentially changes if the external loading is applied. In particular, the manipulator resistance in the $x$-direction becomes lower and lower while the force $F_x$ is increasing (see Fig. 9a). In contrast, the resistance in the $y$-direction with respect to the force $F_y$ becomes higher and higher while this force is increasing (see Fig. 9b). These results are also confirmed by the $K_{xx}$ and $K_{yy}$ plots presented in Fig. 10, which show enormous lost of $x$-direction resistance under the $F_x$ loading (it can be treated as a "quasi-buckling", see Fig. 10a for the stiffness coefficient $K_{xx}$). On the other side, while increasing the force $F_y$, the stiffness coefficient $K_{yy}$ is very small at the beginning, then it is increasing until reaches the maximum value, and then it is decreasing (see Fig. 10b). In this figure, an evolution of the manipulator configuration under the loading with relevant stiffness coefficients $K_{xx}$ and $K_{yy}$ plots (corresponding to the case $a/b$=0.75) are also presented to explain the above mention results from geometrical and physical point of view, which are corresponded to the stiffness coefficient and force relation. There are presented four representative configurations showing shapes of all segments and their position with respect to the joint limits. As follows from them, the observed sudden change of the stiffness (see Figs. 9 and 10) occurs when one of the segment is close to its joint limits, when the equivalent rotational stiffness coefficient is very low. Hence, in practice it is necessary to avoid applying too high loading causing approaching to the joint limits and losing the manipulator stiffness.

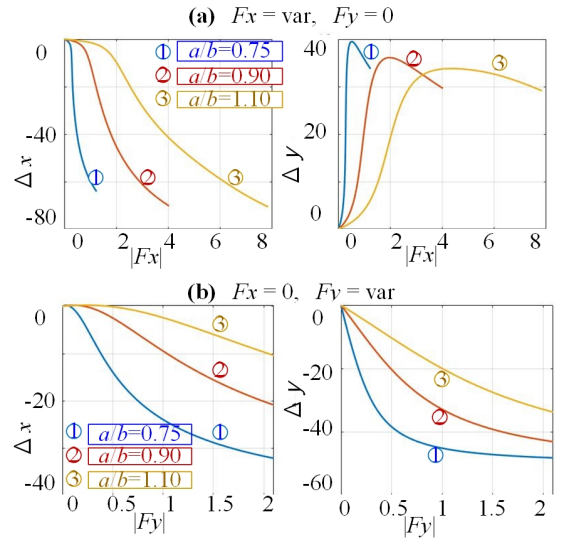

Figure 9. Force-deflection relations of three-segment mechanism for non-straight initial configuration with $(x,y)_o = (5.5b, 0)$.

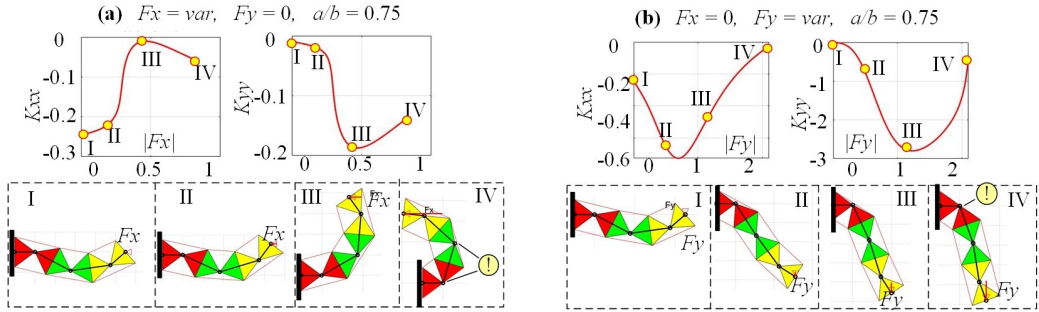

Figure 10. Evolution of the manipulator configuration under the loading.

Therefore, as follows from the above study, mechanical properties of a serial manipulator based on dual-triangle segments have a number of particularities different from a classical serial structure composed of rigid links and compliant components. These particularities must be obligatory taken into account in control algorism ensuring desired motions of such manipulator, which is in the focus of our future research.

## IV. CONCLUSION

The paper focuses on the compliant serial manipulator composed of a new type of dual-triangle tensegrity mechanism, which is composed of rigid triangles connected by passive joints. In contrast to conventional cable driven mechanisms, here there are two length-controllable elastic edges that can generate internal preloading. So, the mechanism can change its equilibrium configuration by adjusting the elastic components initial lengths.

The energy method was used to find the equilibrium configurations for different combination of geometrical and mechanical parameters. The results show that both stable and unstable equilibriums may exist, and the manipulator shape will essential evolute if the external loading is applied. Some analytical results are presented allowing to find the manipulator shape under the loading and to estimate stability of corresponding configuration.

The manipulator stiffness analysis for both loaded and unloaded mode was done using the VJM method, and the relations between the end-effector deflection and the external force were obtained. Similar to the single dual-triangle segment, the buckling phenomenon occurs if the manipulator initial configuration is straight. Besides, for the non-straight initial configuration, the sudden change in deflection was also observed in some cases, which was treated as quasi-buckling. This particularities of the manipulator stiffness behavior was also observed in simulation.

The obtained results allowing to predict manipulator complicated behavior under the loading, and to avoid the buckling or quasi-buckling phenomenon by proper selection of the mechanical parameters, will be used in future for development relevant control algorisms and redundancy resolution.

ACKNOWLEDGMENT

This work was supported by the China Scholarship Council ( No. 201801810036 ).

.

Authors' background (This form is only for submitted manuscript for review)

| Your Name | Title* | Affiliation | Research Field | Personal website |
|---|---|---|---|---|
| Wanda Zhao | Phd candidate | Laboratoire des Sciences du Numérique de Nantes (LS2N), Ecole Centrale Nantes | Robotics, | |
| Anatol Pashkevich | Full Professor | Laboratoire des Sciences du Numérique de Nantes (LS2N), UMR CNRS 6004, IMT Atlantique Nantes | Robotics, Robot Control | https://cv.archives-ouvertes.fr/Anatol-pashkevich |
| Damien Chablat | Full Professor | Laboratoire des Sciences du Numérique de Nantes (LS2N), UMR CNRS 6004, Ecole Centrale Nantes | Robotics, Parallel Manipulator Design, | https://cv.archives-ouvertes.fr/damien-chablat |
| Alexandre Klimchik | Associate Professor | Center Technologies in Robotics and Mechatronics Components, Innopolis University | Robotics, Computer Science | https://www.researchgate.net/profile/Alexandr_Klimchik |

*This form helps us to understand your paper better, the form itself will not be published.

*Title can be chosen from: master student, Phd candidate, assistant professor, lecture, senior lecture, associate professor, full professor